%% file: main.tex
\documentclass{article}
\usepackage{arxiv}
\usepackage{cite}
\usepackage{amsmath,amssymb,amsfonts}
\usepackage{algorithmic}
\usepackage{graphicx}
\usepackage{textcomp}
\usepackage{booktabs}
\usepackage{url}
\usepackage{xcolor}
\usepackage{wrapfig}

\definecolor{trailer}{gray}{0.9}
\newenvironment{trailer}[1]
  {\par\addvspace{12pt}\noindent
   {\fboxsep=0pt\colorbox{trailer}{\parbox{\linewidth}{%
      \vspace{3pt}\hspace{3pt}\textbf{#1\vphantom{Qy}}\vspace{3pt}}}}%
   \par\nopagebreak\vspace{2pt}\noindent\ignorespaces}
  {\par\nopagebreak\noindent\textcolor{trailer}{\rule{\linewidth}{1pt}}%
   \par\addvspace{\baselineskip}}

\begin{document}

\title{Formal Raising: Embedding Verifiable Ethical Properties into the Development of AI}

\author{%
  Rolf Drechsler \qquad Christina Plump\\[0.5ex]
  University of Bremen, Bremen, Germany\\
  DFKI GmbH, Cyber-Physical Systems, Bremen, Germany\\
  \texttt{\{drechsler,cplump\}@uni-bremen.de}%
}

\date{}
%
%
\maketitle

\begin{abstract}Alan Turing proposed constructing a ``child machine'' and educating it rather than directly programming an adult-like intelligent machine. Recent work on \emph{model raising} revives this developmental view for \emph{Large Language Models} by arguing that moral values should accompany capability acquisition from the beginning rather than be added only through post-hoc alignment. This paper proposes \emph{Formal Raising} (FR): selected moral values are represented as explicit, machine-checkable properties that participate in development, evaluation, and action selection. We present a C-based reasoning framework in which candidate actions are evaluated against prioritized ethical properties. Conflicts are resolved w.r.t.\ a given order, strict rules are non-defeasible, and underdetermination is reported explicitly. Exceptions are not encoded as special cases but emerge from the comparison of alternatives. We provide examples that cover conflicts involving life, truthfulness, harm, privacy, autonomy, and strict rules. We outline a property layer inspired by a \emph{Property Specification Language}~(PSL) for state and temporal requirements and discuss its integration into a raising process.
\end{abstract}   


\input{mainContent-arxiv}

\end{document}

%% file: mainContent-arxiv.tex
\section{Introduction}

In his seminal paper \emph{Computing Machinery and Intelligence} from 1950, Turing considered not only the imitation game, but also how an intelligent machine might be obtained. Rather than programming a complete adult mind, he proposed constructing a child-like machine and educating it~\cite{Turing1950}. He separated the problem into a ``child programme'' and an ``education process'', discussed reward and punishment, and considered built-in logical inference, propositions, and imperatives. This is remarkably close to a developmental view of AI.

Modern machine learning techniques and increasing computing capabilities have made Turing's learning component practical at scale, yet the dominant LLM pipeline differs from his educational picture. Models first acquire broad capabilities through pre-training, while behavioral alignment is strengthened later through supervised fine-tuning, \emph{Reinforcement Learning from Human Feedback}~(RLHF), AI feedback, and related mechanisms~\cite{BaiRLHF,BaiConstitutional}. Alignment remains vulnerable: adversarial prompts can circumvent restrictions~\cite{Zou2023}, fine-tuning can degrade safety even without malicious intent~\cite{Qi2024}, and theoretical work identifies limitations of procedures that merely reduce undesired behavior~\cite{Wolf2024}.

Aydin \emph{et al.} therefore recently proposed a change of perspective from \emph{model training} to \emph{model raising}~\cite{Aydin2025}. Their central idea is to move value formation upstream: values should accompany the acquisition of knowledge and capabilities from the earliest stages of development. They propose first-person framing (stating training data as experience in personal form), contextualization as lived experience, social interaction, scaffolded data order, and commitment to values from the first training token onward. This is conceptually close to Turing's child-machine proposal, but adapted to contemporary foundation models.
A further step is now possible because formal verification, temporal logic, property languages, SAT/SMT solving, model checking, runtime verification, and neurosymbolic learning have matured. The \emph{Property Specification Language}~(PSL), for example, captures design intent in a machine-readable form suitable for simulation and formal verification~\cite{PSL2012}, while semantic-based regularization and \emph{Logic Tensor Networks} (LTNs) show that logical constraints can participate in learning and inference~\cite{Diligenti2017,Badreddine2022}. These developments provide the technological basis for combining \emph{model raising} with \emph{formal properties}.

We call the resulting concept \emph{Formal Raising}~(FR). FR is defined with respect to an underlying ethical model, represented through properties, priorities, and invariants, that can be mechanically checked. Specification through a formal model allows inspection, change, or cultural and application-specific adaptation. Formal verification then provides guarantees \emph{relative to the chosen specification and world model}. It is important to note that the concept of Formal Raising does not claim a universally accepted ethical model, as a proof is always with respect to the specified world model and chosen specification.

Our contributions are fourfold. First, we connect Turing's 75-year-old educational idea with model raising and modern formal methods. Second, we define an FR architecture in which ethical properties participate in action selection instead of being used only as post-hoc filters. Third, we present a directly executable C framework that implements prioritized ethical reasoning, including hard constraints, conflict resolution, and explicit detection of underdetermination. Fourth, we outline how temporal/property languages such as PSL can serve as an input layer from which  constraints, training examples, or C-level checks are generated. A secure developmental environment and its influences are discussed in the outlook.


\section{From the Child Machine to Model Raising}
This section traces the developmental view of AI from its origins to its current form. We first revisit Turing's proposal of an educated child machine and then summarize the model-raising approach of Aydin et al., which forms the starting point for Formal Raising.

\subsection{Turing's Developmental View}

Turing's discussion of learning machines is particularly relevant when read from today's perspective. He distinguished three contributors to an adult mind: the initial state, education, and other experience~\cite{Turing1950}. He then asked why one should simulate the adult mind directly instead of constructing a child-like program and subjecting it to education. The resulting problem has two closely coupled components: the structure of the child machine and the educational process.

Turing's proposal contains several modern elements: 
\begin{itemize}
\item the machine changes through interaction,
\item reward and punishment are only part of education,
\item richer symbolic communication and built-in logical inference are considered and 
\item the teacher may not fully understand the machine's internal state. 
\end{itemize}
The last point anticipates a central difficulty of modern deep learning: development can be guided externally while internal representations remain only partially understood.
The key insight for this paper is therefore not that modern AI should imitate children literally. It is that \emph{intelligence can be viewed as the outcome of a structured developmental process}. 

\subsection{Model Raising}

Aydin \emph{et al.} revive this developmental perspective for LLMs~\cite{Aydin2025}. They argue that the separation between capability acquisition and later value alignment has developed historically rather than being a conceptual necessity. Their model-raising proposal aims at early and intrinsic commitment to values by redesigning the training corpus. Instead of treating training data as an unordered collection of detached texts, data should be framed through a persistent first-person perspective, embedded in lived experience, include social interaction, and follow a scaffolded learning order.
Their proposal does not require values to be represented as an explicit formal code; they may be conveyed implicitly through training experience and thus still generate statistically shaped behavior.




\section{Why Formal Raising?}
Model raising leaves open how the values it conveys can be inspected or checked. In this section, we contrast behavioral alignment with explicit formal invariants, argue that the underlying ethical model must be configurable rather than universal, and position FR with respect to existing work on formal machine ethics.
\subsection{From Behavioral Alignment to Explicit Invariants}

Current alignment mechanisms, as introduced by~\cite{BaiConstitutional}, are powerful but predominantly behavioral and statistical. RLHF, for example, optimizes behavior according to preference models~\cite{BaiRLHF}. Constitutional AI makes principles more explicit, but still uses these principles primarily to generate critiques, revisions, preferences, and learning signals~\cite{BaiConstitutional}. FR 
adds a second semantic layer: selected principles are represented as properties with explicit semantics.

\begin{figure}[t]
\centering
\includegraphics[width=0.7\textwidth]{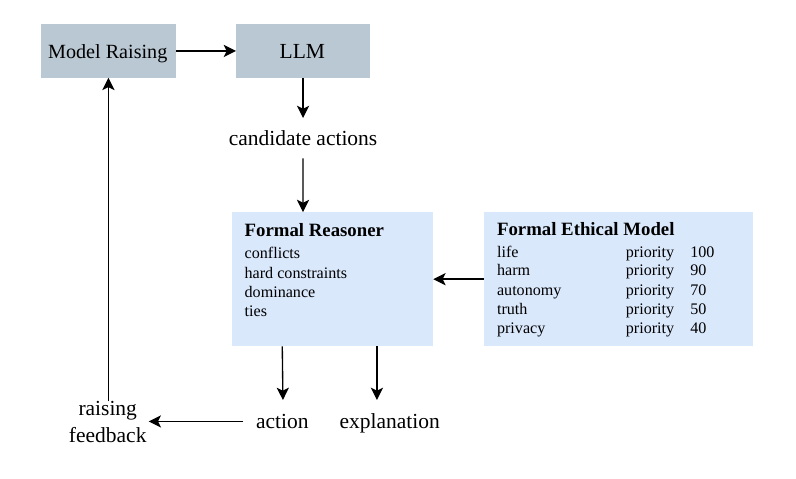}
\caption{Formal Raising: properties participate in development and action selection}\label{fig:architecture}
\end{figure}

The distinction is important: A statistical objective can strongly discourage an undesired behavior while leaving a non-zero probability that it occurs. Instead, a formal invariant states that a particular condition must not be reached within the modeled system. Whether such a guarantee can be established depends on the abstraction, the underlying model, and the verification method. The challenge lies within the identification of parts of the AI-system for which properties can be 
made explicit and checkable.
Figure~\ref{fig:architecture} shows a sketch of the proposed architecture, by expanding Model Raising with a formal reasoner as well as a formal ethical model (light blue boxes for additions by this work).


\subsection{Ethics as a Configurable Model}

A formal ethical model should not be confused with a proof of ``the correct ethics.'' Ethical priorities differ across legal systems, cultures, organizations, and applications. Even within one society, values such as autonomy, privacy, truthfulness, fairness, prevention of harm, and protection of life can conflict. The formalization therefore begins with a declared model
\begin{equation}
\mathcal{E}=\{(P_i,\pi_i,h_i)\mid i=1,\ldots,n\},
\end{equation}
where $P_i$ is an ethical property, $\pi_i$ is its priority, and $h_i$ indicates whether it is a strict ethical rule, and must be treated as a hard constraint by the formal reasoner. The priorities are part of the specification and must remain inspectable.

An explicit ethical model allows disagreement to become explicit. Although two institutions may use the same reasoning engine, different property sets or priorities can lead the reasoning engine to different decisions. Formal verification can determine consequences, expose conflicts and identify cases where the model is insufficient, making ethical debate testable, not unnecessary.

\subsection{Relation to Formal Machine Ethics}

Formal ethical reasoning is not new. Dennis \emph{et al.} proposed and verified ethical plan selection for autonomous systems, ranking available plans according to ethical principles~\cite{Dennis2016}. Their work demonstrates that formal verification can establish that an autonomous agent selects the most ethical plan available according to a specified
policy.

Recently, Priya and Rao proposed a deontic temporal logic for the formal verification of AI ethics~\cite{PriyaRao2026}. Their framework combines deontic concepts such as obligation, permission, and prohibition with temporal operators and uses them to specify and verify system-level ethical properties, including fairness and explainability. The approach is evaluated on COMPAS and loan-prediction systems using automated theorem
proving. This work is particularly relevant to the temporal extension considered later in this paper, since ethical requirements are expressed not only over individual states but also over system behavior over time.
Ghosh and Smith likewise combine deontic logic and formal verification for ethical autonomous systems~\cite{GhoshSmith2026}. They construct model abstractions in which actions are classified as obligatory, permissible, or forbidden, and use temporal-logic specifications and model checking to analyze ethical behavior. Their approach additionally considers changes in environmental context and multi-agent settings and demonstrates the
formalism using an autonomous UAV case study.

These approaches demonstrate that ethical principles can be represented formally and subjected to automated reasoning or verification. FR builds on this general idea but shifts the role of the formal mechanism: It applies ethical verification throughout the entire raising process. Also, a configurable priority structure allows conflict resolution accompanied by explanations.   

\section{Formal Reasoning Model}
We now make the reasoning underlying FR precise. After introducing candidate actions and property violations, we show how conflicts are resolved through a priority ordering and how the model reports situations in which it cannot reach a decision.

\subsection{States, Actions, and Violations}

Let $A=\{a_1,\ldots,a_m\}$ be the set of candidate actions produced by an agent or planner. 
A property $P_i$ is represented by a violation predicate
\begin{equation}
v_i(a)=\begin{cases}
1 & \text{if }a\text{ violates }P_i,\\
0 & \text{otherwise.}
\end{cases}
\end{equation}
The ethical specification contains no action-specific exceptions. This is central to the approach. For instance, 
we do not encode ``lying is allowed when it protects life.'' Instead, protection of life and truthfulness are independent properties with different priorities.

For each priority level $\Pi$, define the number of predicates violated as
\begin{equation}
V_\Pi(a)=\sum_{i:\pi_i=\Pi} v_i(a).
\end{equation}
Actions are compared from the highest priority to the lowest. Given actions $a$ and $b$, let $\Pi^*$ be the highest priority at which $V_{\Pi^*}(a)\neq V_{\Pi^*}(b)$. Then
\begin{equation}
a \prec b \quad \Leftrightarrow \quad V_{\Pi^*}(a)<V_{\Pi^*}(b),
\end{equation}
where $a\prec b$ means that $a$ is ethically preferred to $b$ under the current model. Hard constraints ($P_i$ with $h_i=1$) dominate all defeasible properties, i.e.~an action violating a hard property is rejected whenever an alternative without a hard violation exists.
This scheme is intentionally simple. More sophisticated variants could use partial orders, weighted MaxSAT, defeasible logics, quantitative risk, or probabilistic consequences. 

\subsection{Conflict Resolution through Priority Ordering}

Consider the classical conflict between truthfulness and protection of life in Figure~\ref{fig:lie}. Two candidate actions A and B are evaluated, given a specific scenario. Comparison yields B preferable to A, because property LIFE has a higher priority than property TRUTH.
The ethical model contains no explicit exception to the $P_{\mathrm{truth}}$-property. The exception is a direct consequence of the comparison. The lie remains a violation of truthfulness; it is simply the best available action under the declared priority structure. 
The explanation records the conflict of both independently inspectable properties. Neither property refers to the other, a judgement can only be made after a comparison of violation profiles $V_{\Pi^*}(A)$ and $V_{\Pi^*}(B)$. This modularity has the advantage of preserving information about violated properties, instead of burying this information in an exception.

\begin{figure}[t]
\centering
\begin{minipage}{0.6\columnwidth}
\begin{small}
\begin{verbatim}
Scenario: attacker asks for a person's location

A: Tell the truth               B: Give false information
   LIFE  (100): violated            LIFE  (100): satisfied
   TRUTH  (50): satisfied           TRUTH  (50): violated   

Highest differing priority: 100
Result: B dominates A
\end{verbatim}
\end{small}
\end{minipage}
\caption{An exception to truthfulness arises from priority-induced order of independent properties}
\label{fig:lie}
\end{figure}

\subsection{Underdetermination}

A formal system should explicitly report when it cannot decide. If two candidate actions have identical violation vectors at all priority levels, then the ethical model is underdetermined for that situation. Our framework reports this explicitly. A higher layer may then request human input, introduce additional properties, or choose according to a non-ethical secondary objective. Hiding underdetermination behind an arbitrary tie-breaker would give a false impression of ethical certainty.


\section{Executable C Framework}

To make the proposal concrete, we implemented the core reasoning mechanism in plain C11. C was chosen deliberately: it makes the semantics visible, minimizes dependencies, and demonstrates that the idea does not rely on a particular ML/AI framework or theorem prover. The code below contains the essential data structures, violation predicates, ethical specification, and a comparison algorithm. We provide some insight on the program structure followed by examples demonstrating the features of the tool.
Fig.~\ref{fig:MainFramework} gives the core representation used in the demonstrator:
\begin{figure}[ht]
\centering
\begin{minipage}{0.68\textwidth}
\begin{small}
\begin{verbatim}
#include <stdbool.h>

typedef struct {
  bool deliberate_lie;
  bool disclose_private_information;
  bool cause_serious_harm;
  bool endanger_human_life;
  bool violate_autonomy;
  bool discriminate;
} EthicalState;

typedef struct {
  const char *name;
  EthicalState state;
} Action;

typedef bool (*Predicate)(const EthicalState *s);

typedef struct {
  const char *name;
  int priority;
  bool hard;
  Predicate violation;
} EthicalProperty;

static bool v_life(const EthicalState *s)
{ return s->endanger_human_life; }
static bool v_harm(const EthicalState *s)
{ return s->cause_serious_harm; }
static bool v_truth(const EthicalState *s)
{ return s->deliberate_lie; }
static bool v_privacy(const EthicalState *s)
{ return s->disclose_private_information; }

static const EthicalProperty properties[] = {
 {"Protection of Life",100,false,v_life},
 {"Avoid Serious Harm", 90,false,v_harm},
 {"Non-Discrimination",80,true, v_discrimination},
 {"Respect Autonomy",  70,false,v_autonomy},
 {"Truthfulness",      50,false,v_truth},
 {"Privacy",           40,false,v_privacy}
};
\end{verbatim}
\end{small}
\end{minipage}
\caption{Property Table and Data Structure of C Framework}\label{fig:MainFramework}
\end{figure}

The property table separates ethical policy from the engine. A new property can be added by defining a predicate and adding an entry to the table. The reasoning engine does not need to know what ``truth'', ``privacy'', or ``autonomy'' mean. It only evaluates violation predicates and priorities. This is analogous to the separation of a design from its property specification in formal hardware verification. The core comparison procedure is shown in Fig.~\ref{fig:CompAlg}. 
\begin{figure}
\centering
\begin{minipage}{0.5\textwidth}
\begin{small}
\begin{verbatim}
static int compare_actions(const Action *a,
                           const Action *b)
{
  int levels[MAX_PRIORITIES];
  size_t n, i;
  bool ha = violates_hard_property(a);
  bool hb = violates_hard_property(b);

  if (ha && !hb) return  1;
  if (!ha && hb) return -1;

  n = collect_priorities(levels,MAX_PRIORITIES);
  for (i=0; i<n; ++i) {
    int va=count_violations(a,levels[i]);
    int vb=count_violations(b,levels[i]);
    if (va<vb) return -1;
    if (vb<va) return  1;
  }
  return 0; /* underdetermined */
}
\end{verbatim}
\end{small}
\end{minipage}
\caption{Core comparison algorithm of C Framework. For readability, helper functions that collect priority levels and count violations are omitted; they are straightforward iterations over the property table.}\label{fig:CompAlg}
\end{figure}

\begin{trailer}{Ex.~1: Life Versus Truthfulness}
The first scenario corresponds to Fig.~\ref{fig:lie}. ``Tell the truth'' does not violate truthfulness but predicts danger to human life. ``Give false information'' violates truthfulness but avoids the higher-priority life violation. The engine therefore selects the second action. The explanation names priority 100 as decisive. Importantly, the output can still report that truthfulness was violated. FR thereby distinguishes \emph{justified preference} from \emph{absence of conflict}.
\end{trailer}

\begin{trailer}{Ex.~2: Privacy Versus Prevention of Harm}

A second scenario considers confidential information needed in an emergency. One alternative preserves confidentiality but is predicted to cause serious harm; another discloses private information and prevents that harm. With serious harm at priority 90 and privacy at priority 40, disclosure is preferred. Again, no ``emergency privacy exception'' is encoded. The result follows from independent properties.
This mechanism is attractive for model governance because it keeps the policy modular. A jurisdiction or institution that assigns a different priority to privacy can modify the specification without modifying the reasoner. A domain can also declare a property hard, in which case no lower-level tradeoff may override it (see Example 4).
\end{trailer}

\begin{trailer}{Ex.~3: Ethical Underdetermination}
If two alternatives satisfy exactly the same properties, the framework returns equality and reports that additional information is required. In a raising pipeline, such cases are particularly valuable training signals. They identify regions in which the current ethical curriculum or property set does not distinguish behavior. Reviewing these cases with a teacher or a policy process can lead to new training examples or formal properties. 
\end{trailer}

\begin{trailer}{Ex.~4: Hard Constraints and Inadmissible Actions}
Priorities express ethical preferences, and can be weighed for different actions. Some properties, however, may be considered strict moral rules. They have to be seen as hard constraints. Consider non-discrimination as a hard property. Assume one candidate action uses a discriminatory criterion but avoids a minor inconvenience, while a second candidate is non-discriminatory but causes that inconvenience. A purely ordinal optimization would decide based on priority, potentially trading both effects. The classification as strict rule prevents this. The discriminatory action is inadmissible whenever a candidate without a hard violation exists. If no admissible action exists, the framework reports this and this result can be included in the raising process. 
\end{trailer}

\begin{trailer}{Ex.~5: Autonomy and Manipulation}

A further example illustrates why the vocabulary of properties matters. Consider an assistant that recommends a course of action to a user. Candidate $a_1$ uses misleading emotional pressure to steer the user toward the desired decision. Candidate $a_2$ presents the relevant facts, states a recommendation, and leaves the final decision to the user. The first candidate may violate both \textsc{AUTONOMY} at priority 70 and \textsc{NON-MANIPULATION} at priority 60, whereas the second satisfies both properties. The reasoner therefore prefers $a_2$ without requiring a special rule that distinguishes ``good persuasion'' from ``bad persuasion.'' That distinction is made in the abstract state supplied to the verifier.

This example simultaneously exposes a central boundary of the formal approach. The reasoner can only operate on propositions such as \texttt{violate\_autonomy} or \texttt{manipulate\_person}; it does not by itself establish that these propositions correctly describe a natural-language interaction. The abstraction layer must map a rich real-world situation to the symbolic state. Consequently, a formally correct decision may still be inappropriate if the world model incorrectly classifies manipulative behavior as non-manipulative. The guarantee is therefore conditional on the semantic abstraction. This is analogous to formal hardware verification: a proof is meaningful with respect to the modeled design and assumptions, and errors in the model remain outside the proof obligation.
\end{trailer}


Table~\ref{tab:properties} summarizes an illustrative property hierarchy. The concrete ordering is not presented as a universal ethical truth. Its purpose is to make the semantics of the demonstrator explicit and to show how different kinds of principles can coexist in one executable model.

\begin{table}[t]
\caption{Illustrative properties used by the reasoning framework}
\label{tab:properties}
\centering
\scriptsize
\begin{tabular}{p{0.22\textwidth}rp{0.65\textwidth}}
\toprule
Property & Priority & Intended interpretation\\
\midrule
Protection of life & 100 & Avoid actions creating substantial danger to human life.\\
Avoid serious harm & 90 & Avoid serious physical harm.\\
Non-discrimination & hard & Example of a non-defeasible rule.\\
Respect autonomy & 70 & Preserve informed individual decision making.\\
Non-manipulation & 60 & Avoid willfully distorting another person's judgment.\\
Truthfulness & 50 & Avoid deliberate communication of believed falsehoods.\\
Privacy & 40 & Avoid unjustified disclosure of private information.\\
Commitments & 30 & Respect explicit commitments unless dominated by a higher principle.\\
\bottomrule
\end{tabular}
\end{table}

Please note that so far, priorities only give an order, not a ratio scale: priority 100 is not ``twice as valuable'' as priority 50. The numbers establish an order, making the decisive point the highest priority at which two candidates differ. More advanced versions may use partial or contextual orders while retaining explicit, reproducible semantics.


\section{From Properties to the Raising Process}

\begin{figure}[b]
\centering
\includegraphics[width=0.60\textwidth]{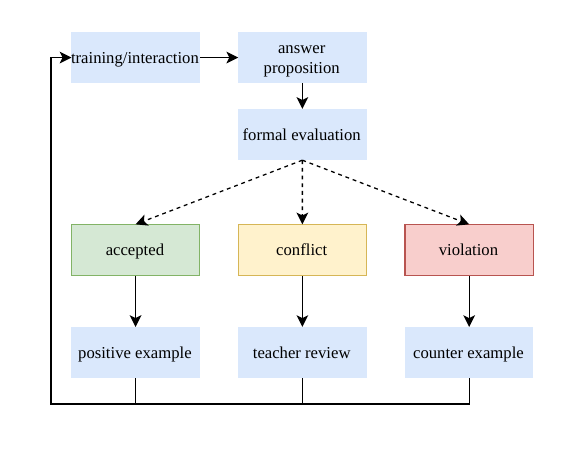}
\caption{Verifier-in-the-raising-loop integration: property outcomes can be converted into examples (for training), feedback (through counterexamples), or requests for clarification from the teacher}\label{fig:loop}
\end{figure}

The C framework demonstrates reasoning at the action level, but FR is intended to influence development and deployment. Fig.~\ref{fig:loop} shows a possible closed loop.


Several integration modes are possible. \emph{Training-data generation} converts formally evaluated situations into positive or negative examples, preference pairs, or explanation traces. \emph{Constraint-based learning} incorporates property satisfaction into the optimization objective, building on work on logical constraints and differentiable logical representations~\cite{Diligenti2017,Badreddine2022}; preserving the crisp meaning of hard properties remains a challenge. \emph{Verifier-guided search} ranks or removes candidate actions before execution and feeds the resulting trace back into later training. \emph{Curriculum generation} turns conflicts and underdetermined cases into progressively harder scenarios, consistent with the scaffolded ordering proposed for model raising~\cite{Aydin2025}.


The C predicates used so far are intentionally elementary. In a larger system, properties should be defined at a higher level and compiled into executable monitors or reasoning constraints. PSL provides an instructive model because it separates property specification from implementation and supports temporal behavior~\cite{PSL2012}. Its design principles can therefore be used to specify an ethical model (see Fig.~\ref{fig:psl}, for example).


\begin{figure}[h]
\centering
\begin{minipage}{0.4\textwidth}
\begin{scriptsize}
\begin{verbatim}
PROPERTY LIFE {
  FORMULA  : never endanger_human_life;
  PRIORITY : 100;
}

PROPERTY TRUTH {
  FORMULA  : never deliberate_lie;
  PRIORITY : 50;
}

PROPERTY PRIVACY {
  FORMULA  : never disclose_private_information;
  PRIORITY : 40;
}
\end{verbatim}
\end{scriptsize}
\end{minipage}
\caption{Illustrative PSL-inspired property input. This is a proposed ethics-oriented front-end, not standard PSL syntax.}
\label{fig:psl}
\end{figure}

Figure~\ref{fig:psl} is a good starting point to illustrate how the underlying behavioral requirement can be expressed using ordinary PSL concepts before adding priorities. In PSL, Boolean expressions can be embedded in temporal operators such as \texttt{always}, \texttt{never}, and \texttt{next}; sequential extended regular expressions can describe finite patterns. For an abstract Boolean signal \texttt{endanger\_human\_life}, a simple invariant can conceptually be written in a PSL style as

\begin{minipage}{0.99\textwidth}
\vspace{0.2cm}
\begin{scriptsize}
\begin{verbatim}
vunit ethical_properties (ethical_model) {
  default clock = (true);
  LIFE: assert always
        (!endanger_human_life);

  TRUTH: assert always
        (!deliberate_lie);

  PRIVACY: assert always
        (!disclose_private_information);
}
\end{verbatim}
\end{scriptsize}
\vspace{0.2cm}
\end{minipage}

Temporal properties become particularly useful for agentic behavior. Many requirements are not properties of one isolated action. Authorization, for example, must occur \emph{before} a protected action; temporary storage of personal information may require later deletion; a detected emergency may require a response within a bounded number of reasoning steps. A PSL-style safety property for authorization can be formulated by maintaining an abstract state proposition \texttt{authorized} and checking

\begin{minipage}{0.99\textwidth}
\vspace{0.1cm}
\begin{scriptsize}
\begin{verbatim}
AUTH: assert always
      (dangerous_action -> authorized);
\end{verbatim}
\end{scriptsize}
\vspace{0.1cm}
\end{minipage}
where the state machine producing \texttt{authorized} guarantees that it becomes true only after valid human authorization. This decomposition is often preferable to inventing a non-standard ``previously'' operator: the protocol state is modeled explicitly and PSL checks the invariant over that state.

For bounded responses, a PSL sequence can express that a trigger is followed by a response within a finite window. An illustrative PSL-style requirement is

\begin{minipage}{0.99\textwidth}
\vspace{0.1cm}
\begin{scriptsize}
\begin{verbatim}
EMERGENCY: assert always
  (life_threat_detected ->
     next_a[1..2](assistance));
\end{verbatim}
\end{scriptsize}
\vspace{0.2cm}
\end{minipage}

meaning that, after a detected threat, assistance must be observed in the specified future window. The precise encoding depends on the chosen PSL flavor.

For FR, temporal logic alone is insufficient, as conflict handling remains unclear. We, therefore, merge the ethics model with PSL-like formulas to an ethics-oriented frontend.

\begin{minipage}{0.99\textwidth}
\vspace{0.2cm}
\begin{scriptsize}
\begin{verbatim}
PROPERTY EMERGENCY_RESPONSE {
  FORMULA  : always
             (life_threat_detected ->
              next_a[1..2](assistance));
  PRIORITY : 100;
  DEFEASIBLE : true;
}

PROPERTY AUTHORIZATION {
  FORMULA  : always
             (dangerous_action -> authorized);
  HARD     : true;
}
\end{verbatim}
\end{scriptsize}
\vspace{0.2cm}
\end{minipage}
This separation is deliberate. PSL supplies a mature conceptual model for specifying temporal behavior, whereas \texttt{PRIORITY}, \texttt{HARD}, \texttt{DEFEASIBLE}, or contextual activation are normative extensions. A compiler can translate the temporal part into a monitor and the metadata into the conflict-resolution structures.

Consider again the privacy example, now as a trace problem. Candidate trace $\tau_1$ detects a life threat, preserves confidentiality, but fails to provide required assistance. Candidate trace $\tau_2$ discloses the minimum required private information and provides assistance in time. The first trace satisfies privacy but violates \textsc{EMERGENCY\_RESPONSE}; the second violates privacy but satisfies the priority-100 emergency property. The same priority-based comparison used for instantaneous actions can therefore be lifted to traces by defining $v_i(\tau)$ as the result of the temporal monitor for property $P_i$.
A generator can translate such declarations to the C property table, runtime monitors, SAT/SMT constraints, or training-data labels. Temporal operators would allow richer requirements, for example, that private information must not be retained after a task, that consent must precede a sensitive action, or that a commitment remains active until it is fulfilled or explicitly released. The step from state predicates to traces is essential for realistic agentic systems because many ethical properties concern sequences rather than isolated actions.

An especially interesting possibility is automated property proposal. An LLM could translate natural-language policy documents into candidate formal properties, while the formal toolchain checks syntax, consistency, redundancy, implication, and satisfiability. Human experts would approve the resulting specification. In this architecture, the LLM helps \emph{author} the model, but only the formally represented properties are used for guarantees. This avoids confusing language-model confidence with formal validity.

\section{Ethical and Security Environment}

Formal Raising also depends on the integrity of the raising environment. Aydin \emph{et al.} emphasize curated interaction and a protected teacher/student setting~\cite{Aydin2025}. Besides the ethical properties considered here, the environment therefore needs \emph{security} properties governing who may modify the curriculum, which tools and data are trusted, whether teacher messages are authentic, and whether external agents can alter the specification. Examples fit the same property-oriented style: ``only authenticated teachers may change priorities'' is a hard access-control property, ``untrusted input must never modify the raising specification'' is an information-flow invariant, and authorization before an external action is a temporal property. A future \emph{Formal Secure Raising} framework could thus combine ethics, safety, and security in one developmental environment.

\section{Discussion and Limitations}

FR makes several assumptions explicit. First, verification is relative to the world model: if consequences are predicted incorrectly, a correct reasoner can still reach the wrong real-world outcome. Second, a total priority order is transparent but potentially too rigid; contextual priorities, partial orders, thresholds, and uncertainty may be required. Third, formalization exposes but cannot settle normative disagreement, for example over privacy versus transparency. A versioned ethical specification can make such choices inspectable rather than leaving them implicit in training data.
Formal properties must influence development rather than becoming another post-hoc governor: evaluation results should affect curriculum selection, data generation, preferences, reinforcement, and model evaluation. Last, scalability remains open. Verifying arbitrary LLM internals is unrealistic with current techniques; a practical approach is compositional, specifying interfaces, abstract states, tool calls, memory operations, and safety-critical sub-decisions while connecting these guarantees to statistical components. Neurosymbolic approaches demonstrate coexistence of symbolic knowledge and neural learning~\cite{Badreddine2022}.

\section{Conclusion}

FR combines the developmental perspective of Turing and model raising with the explicitness of formal methods. Candidate behavior is evaluated against inspectable properties; conflicts are resolved according to declared priorities, hard constraints remain non-defeasible, and underdetermination is explicitly stated. The demonstrator shows that ethical exceptions need not be encoded manually: they can result from the priority structure of independently specified properties, making the conflict transparent in an explanation.

Formal properties cannot determine which ethics society ought to choose, nor can they compensate for an incorrect world model. They can make a chosen ethical model transparent, executable, analyzable, and testable, and feed verified outcomes back into the raising process. Future work includes richer temporal properties, automated property generation, verifier-guided training, multi-agent reasoning, and extension of the same property-oriented approach to security.

\section*{Declaration of AI Usage} {Generative AI models have been used for discussion, feedback and general support. All information has been reviewed by the authors. The paper has been written by the authors with LLM-support on orthography and expression.}